\documentclass[11pt]{article}
\usepackage{acl2016}
\usepackage{times}
\usepackage{amsmath}
\usepackage{latexsym}
\usepackage{graphics}
\usepackage{graphicx}
\usepackage{dblfloatfix}
\usepackage{subfigure}

\usepackage{color}
\usepackage[dvipsnames]{xcolor}

\newcommand{\argmin}{\operatornamewithlimits{argmin}}
\aclfinalcopy % Uncomment this line for the final submission
\title{Joint Embeddings of Hierarchical Categories and Entities}

\author{Yuezhang Li\qquad Ronghuo Zheng \qquad Tian Tian \qquad Zhiting Hu \qquad   Rahul Iyer \qquad Katia Sycara \\
        Carnegie Mellon University\\
	    5000 Forbes Ave.\\
	    Pittsburgh, PA 15213, USA\\
	    {\tt yuezhanl@andrew.cmu.edu}}

\date{}

\begin{document}

\maketitle

\begin{abstract}
Due to the lack of structured knowledge applied in learning distributed representation of categories, existing work cannot incorporate category hierarchies into entity information.~We propose a framework that embeds entities and categories into a semantic space by integrating structured knowledge and taxonomy hierarchy from large knowledge bases. The framework allows to compute meaningful semantic relatedness between entities and categories.~Compared with the previous state of the art, our framework can handle both single-word concepts and multiple-word concepts with superior performance in concept categorization and semantic relatedness.
\end{abstract}

\section{Introduction}

Hierarchies, most commonly represented as Tree or DAG structures, provide a natural way to categorize and locate knowledge in large knowledge bases (KBs).~For example, WordNet, Freebase and Wikipedia use hierarchical taxonomy to organize entities into category hierarchies. These hierarchical categories could benefit applications such as concept categorization \cite{rothenhausler2009unsupervised}, document categorization \cite{gopal2013recursive}, object categorization \cite{verma2012learning}, and link prediction in knowlegde graphs \cite{lin2015learning}.~In all of these applications, it is essential to have a good representation of categories and entities as well as a good semantic relatedness measure.

Existing work does not use structured knowledge of KBs to embed representations of entities and categories into a semantic space.~Current entity embedding methods cannot provide the relatedness measure between entities and categories, although they successfully learn entity representations and relatedness measure between entities \cite{hu2015entity}.~Knowledge graph embedding methods \cite{wang2014knowledge,lin2015learning} give embeddings of entities and relations but lack category representations and relatedness measure.~Though taxonomy embedding methods \cite{weinberger2009large,hwang2014unified} learn category embeddings, they primarily target documents classification not entity representations.

\begin{figure*}[t]
    \centering
    \includegraphics[width=\textwidth]{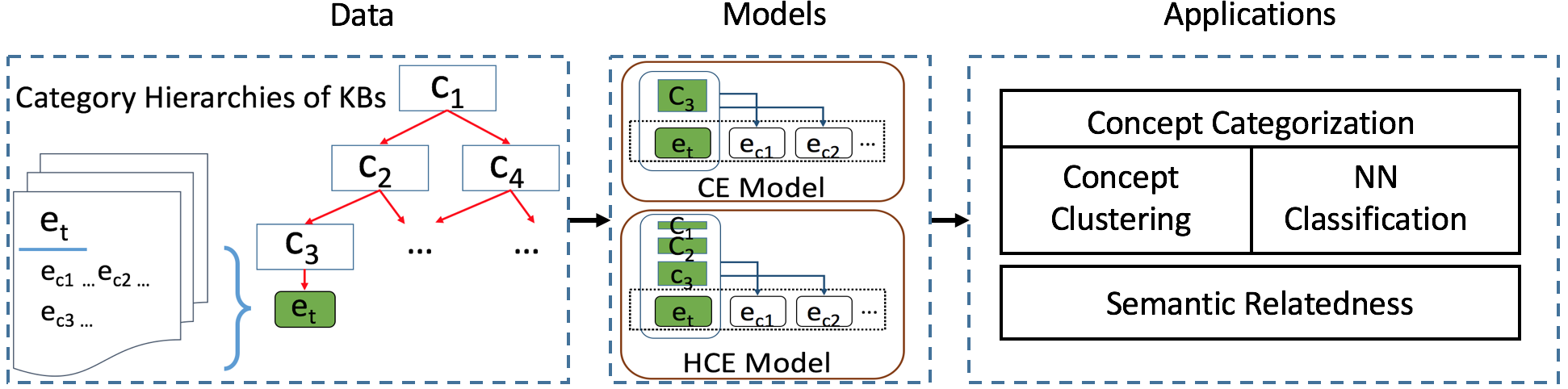}
    \caption{The organization of research elements comprising this paper.}
    \label{fig:1}
\end{figure*}
In this paper, we propose two models to simultaneously learn entity and category vectors from large-scale knowledge bases (KBs).~They are Category Embedding model and Hierarchical Category Embedding model. The {\bf Category Embedding model} (CE model) extends the entity embedding method of \cite{hu2015entity} by using category information with entities to learn entity and category embeddings.~The {\bf Hierarchical Category Embedding model} (HCE model) extends CE model to integrate the hierarchical structure of categories.~It considers all ancestor categories of one entity.~The final learned entity and category vectors can capture meaningful semantic relatedness between entities and categories.

We train the category and entity vectors on Wikipedia, and then evaluate our methods from {\bf two applications: concept categorization} \cite{baroni2010distributional} and {\bf semantic relatedness} \cite{finkelstein2001placing}.

The organization of the research elements that comprise this paper, summarizing the above discussion, is shown in Figure \ref{fig:1}. 

The main contributions of our paper are summarized as follows. First, we incorporate category information into entity embeddings with the proposed CE model to get entity and category embeddings simultaneously. Second, we add category hierarchies to CE model and develop HCE model to enhance the embeddings' quality. Third, we propose a concept categorization method based on nearest neighbor classification that avoids the issues arising from disparity in the granularity of categories that plague traditional clustering methods.~Fourth, our model can handle multiple-word concepts/entities\footnote{In this paper, concepts and entities denote same thing.} such as ``hot dog" and distinguish it from ``dog". Overall, our model outperforms state-of-the-art methods on both concept categorization and semantic relatedness.

% \ks{Can we say something similar for semantic relatedness?}

% \ks{It would be good if we can have a small paragraph summarizing the papaer contributions}

\section{Hierarchical Category Embedding}
In order to find representations for categories and entities that can capture their semantic relatedness, we use existing hierarchical categories and entities labeled with these categories, and explore two methods: 1) {\bf Category Embedding model} (CE Model): it  replaces the entities in the context with their directly labeled categories to build categories' context; 2) {\bf Hierarchical Category Embedding} (HCE Model): it  further incorporates all ancestor categories of the context entities to utilize the hierarchical information.

\subsection{Category Embedding (CE) Model}
Our category embedding (CE) model is based on the Skip-gram word embedding model\cite{mikolov2013distributed}. The skip-gram model aims at generating word representations that are good at predicting \emph{context} words surrounding a \emph{target} word in a sliding window. Previous work \cite{hu2015entity} extends the entity's context to the whole article that describes the entity and acquires a set of entity pairs $\mathbf{D} = \{(e_t, e_c)\}$,  where $e_t$ denotes the {\em target} entity and $e_c$ denotes the {\em context} entity.

Our CE model extends those approaches by incorporating category information.~In KBs such as Wikipedia, category hierarchies are usually given as DAG or tree structures, and entities are categorized into one or more categories as leaves.~Thus, in KBs, each entity $e_t$ is labeled with one or more categories $(c_{1}, c_{2},...,c_{k}), k\geq 1$ and described by an article containing other {\em context} entities (see Data in Figure~\ref{fig:1}).

To learn embeddings of entities and categories simultaneously, we adopt a method that incorporates the labeled categories into the entities when predicting the context entities, similar to TWE-1 model \cite{liu2015topical}. For example, if $e_t$ is the \emph{target} entity in the document, its labeled categories $(c_{1}, c_{2},...,c_{k})$ would be combined with the entity $e_t$ to predict the context entities like $e_{c1}$ and $e_{c2}$ (see CE Model in Figure~\ref{fig:1}).~For each target-context entity pair $(e_t, e_c)$, the probability of $e_c$ being context of $e_t$ is defined as the following softmax:
\begin{equation}
	P(e_c|e_t) = \frac{\exp{(e_t \cdot e_c)}}{\sum_{e \in \mathbf{E}}\exp{(e_t \cdot e)}},
\label{softmax}
\end{equation}
where $\mathbf{E}$ denotes the set of all entity vectors, and $\exp{}$ is the exponential function. For convenience, here we abuse the notation of $e_t$ and $e_c$ to denote a target entity vector and a context entity vector respectively.

Similar to TWE-1 model, We learn the target and context vectors by maximizing the average log probability:
\begin{align}
	L = &\frac{1}{|\mathbf{D}|}\sum_{(e_c,e_t)\in \mathbf{D}} \Big[\log P(e_c|e_t)\notag\\
	&+\sum_{c_i \in \mathbf{C}(e_t)}\log P(e_c|c_i) \Big],
\end{align}
where $\mathbf{D}$ is the set of all entity pairs and we abuse the notation of $c_i$ to denote a category vector, and $\mathbf{C}(e_t)$ denotes the categories of entity $e_t$.

\begin{figure*}[t]
    \centering
    \includegraphics[width = \textwidth]{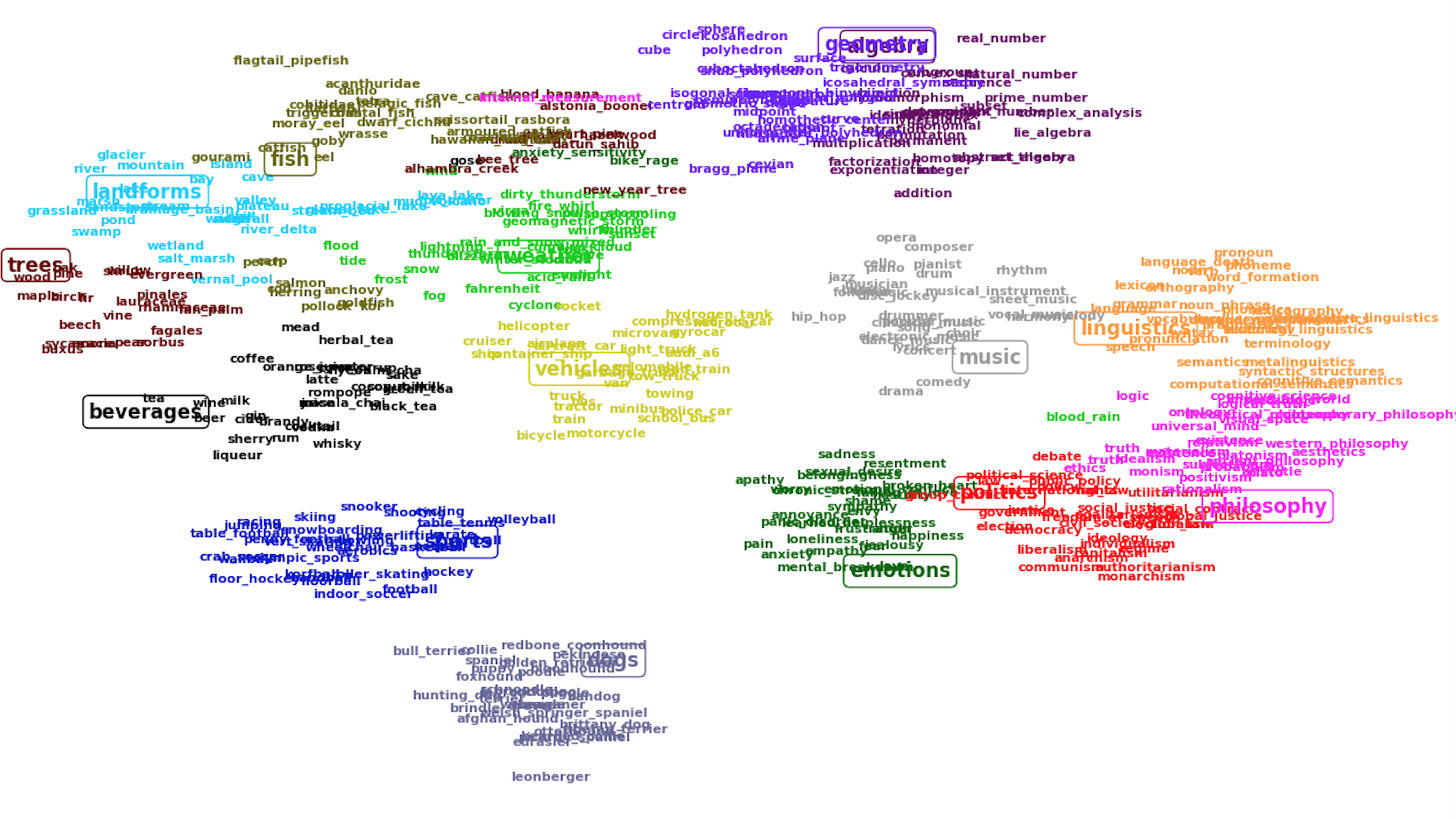}
    \caption{Category and entity embedding visualization of the DOTA-all data set (see Section \ref{ssset:CC_datasets}). We use t-SNE \cite{van2008visualizing} algorithms to map vectors into a 2-dimensional space. Labels with the same color are entities belonging to the same category. Labels surrounded by a box are categories vectors.} 
    \label{vi}
\end{figure*}

\subsection{Hierarchical Category Embedding(HCE) Model}
From the design above, we can get the embeddings of all categories and entities in KBs without capturing the semantics of hierarchical structure of categories. In a category hierarchy, the categories at lower layers will cover fewer but more specific concepts than categories at upper layers. To capture this feature, we extend the CE model to further incorporate the ancestor categories of the target entity when predicting the context entities (see HCE Model in Figure~\ref{fig:1}). If a category is near an entity, its ancestor categories would also be close to that entity. On the other hand,  an increasing distance of the category from the target entity would decrease the power of that category in predicting context entities. Therefore, given the target entity $e_t$ and the context entity $e_c$, the objective is to maximize the following weighted average log probability:
 \begin{align} \label{eq:3}
 	L =& \frac{1}{|\mathbf{D}|}\sum_{(e_c,e_t)\in \mathbf{D}}\Big[\log P(e_c|e_t)\notag\\
 	&+\sum_{c_i \in \mathbf{A}(e_t)}w_i \log P(e_c|c_i)\Big],
 \end{align}
where $\mathbf{A}(e_t)$ represents the set of ancestor categories of entity $e_t$, and $w_i$ is the weight of each category in predicting the context entity. To implement the intuition that a category is more relevant to its closer ancestor, for example, ``NBC Mystery Movies" is more relevant to ``Mystery Movies" than ``Entertainment",  we set $w_i\propto \frac{1}{l(c_c,c_i)}$ where $l(c_c,c_i)$ denotes the average number of steps going down from category $c_i$ to category $c_c$, and it is constrained with $\sum_i w_i = 1$.

Figure \ref{vi} presents the results of our HCE model for DOTA-all data set (see Section \ref{ssset:CC_datasets}). The visualization shows that our embedding method is able to clearly separate entities into distinct categories.

\subsection{Learning}

Learning CE and HCE models follows the optimization scheme of skip-gram model \cite{mikolov2013distributed}. We use negative sampling to reformulate the objective function, which is then optimized through stochastic gradient descent (SGD).

Specifically, the likelihood of each context entity of a target entity is defined with the softmax function in Eq.~\ref{softmax}, which iterates over all entities. Thus, it is computationally intractable. We apply the standard negative sampling technique to transform the objective function in equation (\ref{eq:3}) to equation (\ref{eq:4}) below and then optimize it through SGD:
\begingroup\makeatletter\def\f@size{10}\check@mathfonts
\def\maketag@@@#1{\hbox{\m@th\large\normalfont#1}}%
\begin{align} \label{eq:4}
	L = &\sum_{(e_c,e_t)\in \mathbf{D}}\Big[\log\sigma(e_c\cdot e_t) + \sum_{c_i\in \mathbf{A}(e_t)} w_i \log \sigma(e_c \cdot c_i)\Big] \notag\\&+
	 \sum_{(e_c', e_t)\in \mathbf{D'}}\Big[\log\sigma(-e_c'\cdot e_t) \notag\\&+ \sum_{c_i\in \mathbf{A}(e_t)} w_i \log \sigma(-e_c' \cdot c_i)\Big],
\end{align}
where $\mathbf{D'}$ is the set of negative sample pairs and $\sigma(x)=1/(1+\exp(-x))$ is the sigmoid function.
	
\section{Applications}
We test the quality of our category and entity embedding with two different applications: concept categorization and semantic relatedness.

\subsection{Concept Categorization}
Concept\footnote{In this paper, concept and entity denote the same thing.} categorization, also known as \emph{concept learning} or \emph{noun categorization}, is a process of assigning a concept to one candidate category, given a set of concepts and candidate categories.~Traditionally, concept categorization is achieved by {\bf concept clustering} due to the lack of category representations.~Since our model can generate representations of categories, we propose a new method of using {\bf nearest neighbor (NN) classification} to directly categorize each concept to a certain category.

\subsubsection{Concept Clustering}
The concept clustering is defined as: given a set of single-word concepts like \emph{dog, cat, apple} and the corresponding gold standard categorizations, apply a word space model to project all the concepts to a semantic space and perform clustering. The clustering results can be evaluated by comparing with the gold standard categorizations. 

Previous methods \cite{almuhareb2004attribute,almuhareb2005concept} managed to learn the {\em properties} and {\em attributes} of a concept; for example, the concept {\em dog} has attributes of {\em (dog color)} and {\em (dog size)} and the properties of {\em(dog brown)} and {\em (dog small)}.~By representing these attributes and properties in a high-dimensional space, one can cluster concepts based on common attributes and properties.

\subsubsection{Nearest Neighbor (NN) Classification}
Although the previous methods described above can generate vectors carefully designed for capturing relations and attributes, the number of vector dimensions can be very large for some methods: from 10,000+ to 1,000,000+, e.g., \cite{almuhareb2005concept,rothenhausler2009unsupervised}.~Due to the large dimensionality, the applicable clustering methods are restricted to the ones that can scale to such high dimensions\footnote{\cite{rothenhausler2009unsupervised} use CLUTO \cite{karypis2002cluto}, a clustering toolkit optimized to cluster large-scale data in reasonable time, as their standard measurements. }.~Other methods such as word embedding~\cite{mikolov2013distributed} may need lower dimensionality vectors but suffer from \emph{granularity problems}. Therefore, we propose an alternative method, namely nearest neighbor (NN) classification, and evaluate comparative trade-offs (see Table~\ref{cat_result}).

Using NN classification, we categorize concepts by directly comparing concept vectors with candidate category vectors.~Precisely, given a set of concepts $\mathbf{E}$ and a set of candidate categories $\mathbf{C}$, we convert all concepts to concept vectors and all candidate categories to category vectors.~Then we use the equation $c = \argmin_{c_i \in \mathbf{C}}||c_i - e||$ to assign the concept vector $e$ with category $c$. Note that in this paper, concept and entity denote the same thing so concept vector is exactly the same as entity vector.

\subsubsection{Evaluation Metrics}
Since purity works as a standard evaluation metric for clustering \cite{rothenhausler2009unsupervised}, to compare our model with the concept clustering, we also use purity to measure our model's performance.~Generally, \textbf{purity} is defined as:
\begin{equation} \label{eq:purity}
\text{purity}(\mathbf{\Omega},\mathbf{G})=\frac{1}{n}\sum_{k}{\max_{j}{|\omega_{k} \cap g_j}|},
\end{equation}
where $\mathbf{\Omega}$ denotes a clustering solution of $n$ clusters, $\mathbf{G}$ is a set of gold standard classes, $\omega_k$ represents the set of labels in a cluster and $g_j$ is the set of labels in a class.~A higher purity indicates better model performance.

\subsection{Semantic Relatedness}
We also evaluate the entity and category embeddings by {\em semantic relatedness}. Semantic relatedness measure is a process of assigning one relatedness score for one word pair. We use a set of standard semantic benchmarks.~Those benchmarks consist of word pairs that have manually rated scores 0-10 for semantic relatedness.~The model performance is assessed by calculating the correlation between scores generated by the model and the average scores given by human subjects.

We use the Spearman's rank correlation coefficient \cite{pirie1988spearman} of human assigned scores and system assigned scores to evaluate our result. Note that scores given by humans are not necessarily the {\em gold standard} because of the differences among humans and the difficulty of giving a clear definition of word similarity. However, a good system should agree with humans to some extent and thus should have a relatively high correlation coefficient (Although when the coefficient is high enough, a higher coefficient would not necessarily reflect the better model). 
The Spearman correlation score is defined by the Pearson correlation coefficient between the ranked variables. For scores $X_i$, $Y_i$ that are in sample size $n$, we use $x_i$ and $y_i$ to denote the rankings of scores $X_i$, $Y_i$. The Spearman correlation coefficient is defined as:
\begin{equation}
	\rho = 1 - \frac{6\sum_{i=1}^n(x_i-y_i)^2}{n(n^2-1)}.
\end{equation}

\section{Experiments}
In the experiments, we use the dataset collected from Wikipedia on Dec.~1, 2015\footnote{https://dumps.wikimedia.org/wikidatawiki/20151201/} as the training data. We preprocess the category hierarchy by pruning administrative categories and deleting bottom-up edges to construct a DAG. The final version of data contains 5,373,165 entities and 793,856 categories organized as a DAG with a maximum depth of 18. The root category is ``main topic classifications".~We train category and entity vectors in various dimensions of 100, 200, 250, 300, 400, 500, with batch size $B=500$ and negative sample size $k=10$.

With the training dataset defined above, we conduct experiments on two applications:~concept categorization and semantic relatedness.

\subsection{Concept Categorization}
In this section, we first introduce datasets applied in concept categorization, and then show baselines followed by experimental results.

\subsubsection{Datasets}\label{ssset:CC_datasets}
There are two datasets used in this experiment. The first one is the {\bf Battig} test set introduced by \cite{baroni2010distributional}, which includes 83 concepts from 10 categories. The {\bf Battig} test set only contains single-word concepts without any multiple-word concepts (e.g., ``table tennis"). Hence, using this dataset restricts the power of concept categorization to single-word level. We use this dataset because it has been used as a benchmark for most previous approaches for concept categorization.

Due to the limitations of the {\bf Battig} test set, we construct a new entity categorization dataset {\bf DOTA }(Dataset Of enTity cAtegorization) with 450 entities categorized into 15 categories (refer to Appendix A).~All the categories and entities are extracted from Wikipedia, so the resulting dataset does not necessarily contains only single-word entities.~Thus, the dataset can be split into two parts, {\bf DOTA-single} that contains 300 single-word entities categorized into 15 categories and {\bf DOTA-mult} that contains 150 multiple-word entities categorized into the same 15 categories.~We design the {\bf DOTA} dataset based on the following principles:
\begin{itemize}
\item {\bf Coverage vs Granularity:} Firstly, the dataset should cover at least one category of Wikipedia's main topics including ``Culture", ``Geography", ``Health", ``Mathematics", ``Nature", ``People", ``Philosophy", ``Religion", ``Society" and ``Technology".~Secondly, categories should be in different granularity, from large categories (e.g.,``philosophy") to small categories (e.g., ``dogs").~Large categories are ones that are located within 5 layers away from the root, medium categories are 6-10 layers away from the root, while small categories have distance of 11-18 to the root.~Our dataset consists of 1/3 large categories, 1/3 medium categories, and 1/3 small categories.
\item {\bf Single-Words vs Multiple-Words:} Previous concept categorization datasets only contain single-words.~However, some concepts are multiple-words and cannot be simply represented by single-words.~For example, the concept ``hot dog" is very different from the concept ``dog".~Word-level embedding cannot solve this problem without phrase recognition, while entity-level embedding can solve it naturally.~Therefore, we make each category of the dataset contain 10 multiple-word entities and 20 single-word entities.
\end{itemize}

\subsubsection{Baselines}\label{subsubsec:CC_baselines}
{\bf Word Embedding (WE)} trained with neutral networks on large corpus provides a way to map a given text to a semantic space. We compare our entity and category embeddings with two word embeddings.
\begin{itemize}
	\item {\bf WE$_{Mikolov}$}\cite{mikolov2013distributed}: \cite{baroni2014don} conducted thorough experiments on word counts and predictive based methods on word representation.~Their experimental results show that Mikolov's word embedding achieves state-of-the-art results in concept categorization.~We trained word embeddings with Mikolov's word2vec toolkit\footnote{https://code.google.com/archive/p/word2vec/} on the same Wikipedia corpus as ours (1.7 million tokens) and then applied the Skip-gram model with negative sample size of 10 and window size of 5 to vector dimensionality of 100, 200, 250, 300, 400, 500 respectively. The best results of various parameter settings are reported in Table \ref{purity1} and Table \ref{purity2}.
	\item {\bf WE$_{Senna}$} \cite{collobert2011natural}: We downloaded this 50-dimension word embedding\footnote{http://ronan.collobert.com/senna/} trained on Wikipedia over 2 months.~We use this embedding as a baseline because it is also trained on Wikipedia.
\end{itemize}

To evaluate the advantage of utilizing category hierarchy in training entity and category embedding, we also compare our 
Hierarchical Category Embedding (HCE) model with our Category Embedding (CE) model that has no hierarchical information. 

\subsubsection{Results}\label{subsubsec:CC_result}

In the experiments, we used scikit-learn \cite{pedregosa2011scikit} to perform clustering. We tested k-means and hierarchical clustering with different distance metrics (euclidean, cosine) and linkage criterion (ward, complete, average). We reported the best result across different clustering parameters in each experiment.

Table.~\ref{purity1} shows the experimental results of the {\bf concept clustering} method.~It is clear that {\bf hierarchical category embedding} (HCE) model outperforms other methods in all datasets.~For single-word entity categorization in Battig and DOTA-single, our HCE model gives a purity of 89\% and 92\% respectively; while for multiple-word entity categorization in DOTA-mult and DOTA-all, the corresponding purities are 91\% and 89\% (higher is better).

\begin{table}[h]
\LARGE
\centering
\resizebox{\columnwidth}{!}{
\begin{tabular}{|l|l|l|l|l|}
\hline
               & {\bf Battig} & {\bf DOTA-single} & {\bf DOTA-mult}	& {\bf DOTA-all}\\ \hline
WE$_{Senna}$   &  0.74   &   0.61   & 0.43 	& 0.45       \\ \hline
WE$_{Mikolov}$ &  0.86   &   0.83   & 0.73   & 	0.78 		\\ \hline
CE (Ours)      &  0.84   &   0.86   &   0.83  	&	0.85\\ \hline
HCE (Ours)     &  {\bf0.89}  & {\bf 0.92}    & {\bf 0.88}	&	{\bf 0.89}		\\ \hline
\end{tabular}
}
\caption{Purity of {\bf concept clustering} method with different embeddings on Battig and DOTA datasets. The numbers given are the best scores achieved under different settings.}
\label{purity1}
\end{table}

The excellent performance of our HCE model on DOTA-mult and DOTA-all lies in the fact that it can naturally produce multiple-word entities.~Since Mikolov's word embeddings can also capture common phrases, it performs well on DOTA-mult dataset with a purity of 73\%. However, the Senna word embeddings only contain single-words.~To get the embeddings of multiple-word, we use the mean word vectors to denote multiple-word embeddings.~As the meaning of a multiple-word is not simply the aggregation of the meaning of the words it contains, the purity of using Senna drops dramatically from 61\% on DOTA-single to 43\% only on DOTA-mult.

Table.\ref{purity2} shows the experimental results of the {\bf nearest neighbor (NN) classification} method. The results indicate the feasibility of using category vectors to directly predict the concept categories without clustering entities. By changing the concept categorization method from {\bf concept clustering} to {\bf nearest neighbor classification}, our model still achieves a purity of 87\% on Battig and around 90\% on DOTA.

\begin{table}[h]
\LARGE
\centering
\resizebox{\columnwidth}{!}{
\begin{tabular}{|l|l|l|l|l|}
\hline
               & {\bf Battig} & {\bf DOTA-single} & {\bf DOTA-mult}	& {\bf DOTA-all}\\ \hline
WE$_{Senna}$   &  0.44  &   0.62   & 0.32 	& 0.33       \\ \hline
WE$_{Mikolov}$ &  0.74  &   0.74   & 0.67   & 0.69 		\\ \hline
CE (Ours)      &  0.79  &   0.89   & 0.85  	& 0.88\\ \hline
HCE (Ours)     &  {\bf 0.87}  & {\bf 0.93}    & {\bf 0.91}	&	{\bf 0.91}		\\ \hline
\end{tabular}
}
\caption{Purity of {\bf nearest neighbor (NN) classification} method with different embeddings on Battig and DOTA datasets. The numbers given are the best scores achieved under different settings.}
\label{purity2}
\end{table}

Table.\ref{cat_result} presents the best prediction results produced by our HCE model with two different methods, namely concept clustering and nearest neighbor (NN) classification.

\begin{table*}[t]
\centering
\small
\begin{tabular}{|l|p{6.5cm}|p{6.5cm}|}
\hline
{\bf Category}   & {\bf Misclassified entities using clustering}   &                        {\bf Misclassified entities using NN classifier} \\ \hline
beverages          &      pear    &   pear                                                                                                                           \\ \hline
sports     & -    &   -                                                         \\ \hline
emotions   &    - &     jumping                             \\ \hline
weather    &  -   &    racing, snowboarding, ship, airplane, rocket, wetland, koi, tetra,  flatfish, whisky                                                                            \\ \hline
landforms  &   flood, sandstone             &   perch	                                                                                             \\ \hline
trees      & mead  &      -                                                                                                                     \\ \hline
algebra    & trigonometry, circle, square, polyhedron, surface, sphere, cube, icosahedron, hemipolyhedron, digon, midpoint, centroid, octadecagon, curvature, curve, zonohedron, cevian, orthant, cuboctahedron, midsphere & curve, curvature 
\\ \hline
geometry   & -&multiplication
                               \\ \hline
fish       & -      &        -                                                                        \\ \hline
dogs       &  -  &    cider, cod                                  
\\ \hline
music      & - &            -                                                                                                            \\ \hline
politics   & idealism, ethics &  idealism, ethics
                             \\ \hline
philosophy &ideology & ideology
                    \\ \hline
linguistics & - &-\\ \hline

vehicles   &  -      &      cycling                                                                                           \\ \hline
\end{tabular}
\caption{Best prediction results of HCE by concept clustering and  nearest neighbor classification on DOTA-single dataset.}
\label{cat_result}
\end{table*}

As for the concept clustering approach, the general performance is very good except for one extreme case that many entities of \emph{geometry} category are misclassified into \emph{algebra} category. Refer to Figure~\ref{vi}, it is clear that because \emph{algebra} category overlaps with \emph{geometry} category, the concept clustering clusters them together. This phenomenon is caused by the difference in granularity of categories.
 
As for the nearest neighbor (NN) classification approach, we found an interesting phenomenon -- several entities from other categories are misclassified into \emph{weather} category.~Referring to the work of \cite{radovanovic2010hubs,marcobaroni2015hubness}, this phenomenon is called \emph{hubness}, which depicts that some vectors (``hubs") tend to appear in the top neighbor lists of many test items in high-dimensional space. In our case, the category vector ``weather" tend to be a ``hub" vector.

Based on the analysis of these two methods above, we can conclude that the NN classifier can address granularity problem but suffers from the \emph{hubness} problem, while concept clustering can deal with \emph{hubness} but has the granularity problem.

\subsection{Semantic relatedness}
We now introduce the datesets and baselines for measuring semantic relatedness and show the experimental results.
\subsubsection{Datasets}
We use a set of standard datasets and preprocess them to fit our method.~Our method requires mapping the words to corresponding Wikipedia entities or categories. For example, we map the word ``cat" to the Wikipedia entity ``cat" and the word ``equipment" to the Wikipedia category ``equipment".~Without loss of generality, we first match a word to a Wikipedia entity based on lexical similarity. If there is no matched entity, we match it to a Wikipedia category. However, it is difficult for this approah to map some words to a specific Wikipedia entity or category because of two reasons:
\begin{itemize}
	\item Wikipedia is a knowledge base that organizes categories and concepts, but some words like adjectives (e.g. smart/stupid) cannot be mapped to any Wikipedia entity or category. Moreover, our entity based approach cannot capture word pairs with lexical differences such as ``swim/swimming".~We thus eliminate these kinds of words.
	\item Some words are ambiguous so they have multiple corresponding entities in Wikipedia. For example, the word ``doctor" can work as entity ``Doctor(title)" that means the holder of an accredited doctoral graduate degree, and it can also be entity ``Physician" that means a professional who practices medicine.~Therefore, we simply discard all these ambiguous words.
\end{itemize}
 Using the filtered datasets does not affect the fairness of our comparison, since we conduct all the other baselines on the same subsets.

{\bf WS:} The WordSim353 dataset is introduced by \cite{finkelstein2001placing}.~It contains 353 pairs of words and their semantic relatedness scores assigned by 13 to 16 human subjects. The work of \cite{agirre2009study} splits the WS-353 dataset into two separate subsets: similarity subset ({\bf WSS}) and relatedness subset ({\bf WSR}).~The former one contains tighter taxonomy relations (e.g., plane/car, student/professor) whereas the latter contains tighter topical relations (e.g., Jerusalem/Israel, OPEC/country). After data preprocessing, 184 pairs of words remain in WS-353, 105 pairs in WSS-353 (202 pairs originally), and 125 pairs in WSR-353 (251 pairs originally).

{\bf MEN:} The work of \cite{bruni2014multimodal} constructed this dataset with 3000 word pairs that have semantic relatedness scores obtained by crowd-sourcing. After data preprocessing, there are 1508 pairs of words left. 

{\bf RG:} A classic dataset contains 65 word pairs introduced by \cite{rubenstein1965contextual}. After data preprocessing, 28 pairs of words are left. 

\subsubsection{Baselines}
We compare our methods with some state-of-the-art methods below.

{\bf WN+NR:} In \cite{radhakrishnan2013extracting}, word similarity measure is derived from Wikipedia category names integrated with WordNet similarity measure by performing regression using a Support Vector Machine. {\bf WN+NR$_1$} and {\bf WN+NR$_2$} are two of the best models reported in their paper.

{\bf WE$_{Mikolov}$}:~As described in Section~\ref{subsubsec:CC_baselines}, we trained word embedding using Mikolov's word2vec toolkit{\footnote{https://code.google.com/archive/p/word2vec/} on the same Wikipedia corpus (1.7 million tokens) as ours to make them comparable. We use Skip-gram model with negative sample size of 10 and window size of 5 , with vector dimensionality of 100, 200, 250, 300, 400, 500. We report the best results obtained from various parameter settings in Table~\ref{wordsim-table}.

{\bf WE$_{Senna}$} \cite{collobert2011natural}: We downloaded this 50-dimension word embedding\footnote{http://ronan.collobert.com/senna/} trained on Wikipedia over 2 months.

\subsubsection{Results}
Table~\ref{wordsim-table} shows the experimental results of the semantic relatedness tasks. We can see that our HCE model yields best results of 57\%, 69\%, and 83\% on WS, MEN and RG datasets. This performance suggests that entity and category embeddings can be used as an indicator of semantic relatedness between words. For WSS dataset, the result of our method is comparable with WN+NR$_1$ method, which integrates WordNet similarity measures with the normalized representation of category names. We also found that Mikolov's word embedding performs better than our method on WSR dataset, but performs worse than our method on WSS dataset. The reason may be that the WSR dataset concentrates on topical related words rather than taxonomy related words, and our method can better capture taxonomy relationship than topic relationship. 

\begin{table}[h]
\small
\centering
\resizebox{\columnwidth}{!}{
\begin{tabular}{|l|l|l|l|l|l|}
\hline
               & \multicolumn{1}{l|}{WS}   & \multicolumn{1}{l|}{WSS}  & \multicolumn{1}{l|}{WSR}  & \multicolumn{1}{l|}{MEN} & \multicolumn{1}{l|}{RG}\\ \hline
WN+NR$_1$      & \multicolumn{1}{l|}{0.55}     & \multicolumn{1}{l|}{{\bf 0.67}}     & \multicolumn{1}{l|}{0.46}     & \multicolumn{1}{l|}{0.53}    & \multicolumn{1}{l|}{0.40}\\ \hline
WN+NR$_2$      & \multicolumn{1}{l|}{0.43}     & \multicolumn{1}{l|}{0.53}     & \multicolumn{1}{l|}{0.32}     & \multicolumn{1}{l|}{0.47}    & \multicolumn{1}{l|}{0.49} \\ \hline
WE$_{Mikolov}$ & \multicolumn{1}{l|}{0.55}     & \multicolumn{1}{l|}{0.64}     & \multicolumn{1}{l|}{{\bf 0.59}}     & \multicolumn{1}{l|}{0.66}    & \multicolumn{1}{l|}{0.75}\\ \hline
WE$_{Senna}$   & \multicolumn{1}{l|}{0.47} & \multicolumn{1}{l|}{0.59} & \multicolumn{1}{l|}{0.38} & \multicolumn{1}{l|}{0.57}    & \multicolumn{1}{l|}{0.47}   \\ \hline
CE (Ours)      & 0.55                         &  0.63                         &       0.49                    &    0.64             & 0.78 \\\hline
HCE (Ours)     & {\bf 0.57}                      & {\bf 0.67}                      & 0.53                      &           {\bf 0.69}                 & {\bf 0.83}                        \\ \hline\end{tabular}
}
\caption{Correlation with human judgements on semantic relatedness evaluation datasets.}
\label{wordsim-table}
\end{table}

\section{Conclusion}
In this paper, we proposed a framework to learn entity and category embeddings to capture semantic relatedness between entities and categories. This framework can incorporate taxonomy hierarchy from large scale knowledge bases. Experiments on both concept categorization and semantic relatedness show that our approach outperforms state of the art approaches. In the future work, we aim at applying our method to more applications such as hierarchical document classification.

\bibliography{sample}
\bibliographystyle{acl2016}

\newpage
\onecolumn
\appendix
\section{The DOTA dataset: 300 single-word entities and 150 multi-word entities from 15 Wikipedia Categories}
\begin{table*}[h]
\small
\centering
\label{my-label}
\begin{tabular}{|l|p{14cm}|}
\hline
\textbf{Category}   & \textbf{Entities}                                                                                                                                                                                                                                                                                                                                                                                              \\ \hline
beverages  & juice, beer, milk, coffee, tea, cocktail, wine, liqueur, sake, vodka, mead, sherry, brandy, gin, rum, latte, whisky, cider, gose, rompope, orange juice, masala chai, green tea, black tea, herbal tea, coconut milk, corn syrup, soy milk, rose water, hyeonmi cha                                                                                                                                                             \\ \hline
sports     & bowling, football, aerobics, hockey, karate, korfball, handball, floorball, skiing, cycling, racing, softball, shooting, netball, snooker, powerlifting, jumping, wallball, volleyball, snowboarding, table tennis, floor hockey, olympic sports, wheelchair basketball, crab soccer, indoor soccer, table football, roller skating, vert skating, penny football                                                                  \\ \hline
emotions   & love, anxiety, empathy, fear, envy, loneliness, shame, anger, annoyance, happiness, jealousy, apathy, resentment, frustration, belongingness, sympathy, pain, worry, hostility, sadness, broken heart, panic disorder, sexual desire, falling in love, emotional conflict, learned helplessness, chronic stress, anxiety sensitivity, mental breakdown, bike rage                                                                  \\ \hline
weather    & cloud, wind, thunderstorm, fog, snow, wave, blizzard, sunlight, tide, virga, lightning, cyclone, whirlwind, sunset, dust, frost, flood, thunder, supercooling, fahrenheit, acid rain, rain and snow mixed, cumulus cloud, winter storm, blowing snow, geomagnetic storm, blood rain, fire whirl, pulse storm, dirty thunderstorm                                                                                                  \\ \hline
landforms  & lake, waterfall, stream, river, wetland, marsh, valley, pond, sandstone, mountain, cave, swamp, ridge, plateau, cliff, grassland, glacier, hill, bay, island, glacial lake, drainage basin, river delta, stream bed, vernal pool, salt marsh, proglacial lake, mud volcano, pit crater, lava lake                                                                                                                                  \\ \hline
trees      & wood, oak, pine, evergreen, willow, vine, shrub, birch, beech, maple, pear, fir, pinales, lauraceae, sorbus, buxus, acacia, rhamnaceae, fagales, sycamore, alhambra creek, alstonia boonei, atlantic hazelwood, bee tree, blood banana, datun sahib, druid oak, new year tree, heart pine, fan palm                                                                                                                                \\ \hline
algebra    & addition, multiplication, exponentiation, tetration, polynomial, calculus, permutation, subgroup, integer, monomial, bijection, homomorphism, determinant, sequence, permanent, homotopy, subset, factorization, associativity, commutativity, real number, abstract algebra, convex set, prime number, complex analysis, natural number, complex number, lie algebra, identity matrix, set theory
\\ \hline
geometry   & trigonometry, circle, square, polyhedron, surface, sphere, cube, icosahedron, hemipolyhedron, digon, midpoint, centroid, octadecagon, curvature, curve, zonohedron, cevian, orthant, cuboctahedron, midsphere, regular polygon, uniform star polyhedron, isogonal figure, icosahedral symmetry, hexagonal bipyramid, snub polyhedron, homothetic center, geometric shape, bragg plane, affine plane
                               \\ \hline
fish       & goldfish, gourami, koi, cobitidae, tetra, goby, danio, wrasse, acanthuridae, anchovy, carp, catfish, cod, eel, flatfish, perch, pollock, salmon, triggerfish, herring, cave catfish, coachwhip ray, dwarf cichlid, moray eel, coastal fish, scissortail rasbora, flagtail pipefish, armoured catfish, hawaiian flagtail, pelagic fish                                                                                            \\ \hline
dogs       & spaniel, foxhound, bloodhound, beagle, pekingese, weimaraner, collie, terrier, poodle, puppy, otterhound, labradoodle, puggle, eurasier, drever, brindle, schnoodle, bandog, leonberger, cockapoo, golden retriever, tibetan terrier, bull terrier, welsh springer spaniel, hunting dog, bearded collie, picardy spaniel, afghan hound, brittany dog, redbone coonhound                                                           \\ \hline
music      & jazz, blues, song, choir, opera, rhythm, lyrics, melody, harmony, concert, comedy, violin, drum, piano, drama, cello, composer, musician, drummer, pianist, hip hop, classical music, electronic music, folk music, dance music, musical instrument, disc jockey, popular music, sheet music, vocal music                                                                                                                          \\ \hline
politics   & democracy, law, government, liberalism, justice, policy, rights, utilitarianism, election, capitalism, ideology, egalitarianism, debate, regime, globalism, authoritarianism, monarchism, anarchism, communism, individualism, freedom of speech, political science, public policy, civil society, international law, social contract, election law, social justice, global justice, group conflict                                \\ \hline
philosophy & ethics, logic, ontology, aristotle, plato, rationalism, platonism, relativism, existence, truth, positivism, metalogic, subjectivism, idealism, materialism, aesthetics, probabilism, monism, truth, existence, western philosophy, contemporary philosophy, cognitive science, logical truth, ancient philosophy, universal mind, visual space, impossible world, theoretical philosophy, internal measurement
                    \\ \hline
linguistics & syntax, grammar, semantics, lexicon, speech, phonetics, vocabulary, phoneme, lexicography, language, pragmatics, orthography, terminology, pronoun, noun, verb, pronunciation, lexicology, metalinguistics, paleolinguistics, language death, historical linguistics, dependency grammar, noun phrase, comparative linguistics, word formation, cognitive semantics, syntactic structures, auxiliary verb, computational semantics \\ \hline
vehicles   & truck, car, aircraft, minibus, motorcycle, microvan, bicycle, tractor, microcar, van, ship, helicopter, airplane, towing, velomobile, rocket, train, bus, gyrocar, cruiser, container ship, school bus, road train, tow truck, audi a6, garbage truck, hydrogen tank, light truck, compressed air car, police car                                                                                                                  \\ \hline

\end{tabular}
\end{table*}

\end{document}